\documentclass[review]{elsarticle}

\usepackage{lineno,hyperref}
\usepackage{adjustbox}
\usepackage{multirow}
\usepackage{multicol, blindtext}
\modulolinenumbers[5]

\journal{Journal of \LaTeX\ Templates}

\biboptions{authoryear}

\begin{document}

\begin{frontmatter}

\title{A Multi-Agent System for Solving the Dynamic Capacitated Vehicle Routing Problem with Stochastic Customers using Trajectory Data Mining}



\author[addressUFMG]{Juan Camilo Fonseca-Galindo}
\ead{juankmilofg@ufmg.br}

\author[addressUNICAMP]{Gabriela de Castro Surita}
\ead{g139095@dac.unicamp.br}

\author[addressUFMG]{José Maia Neto}
\ead{jmnt@ufmg.br}

\author[addressUFMG]{Cristiano Leite de Castro}
\ead{crislcastro@ufmg.br}

\author[addressUFMG]{André Paim Lemos}
\ead{andrepaim@ufmg.br}

\address[addressUFMG]{Graduate Program in Electrical Engineering, Federal University of Minas Gerais, Av. Antônio Carlos 6627, 31270-901, Belo Horizonte, MG, Brazil}

\address[addressUNICAMP]{Graduate Program in Electrical Engineering, University of Campinas, Cidade Universitária Zeferino Vaz, Barão Geraldo, Campinas, SP, Brazil}

\begin{abstract}
The worldwide growth of e-commerce has created new challenges for logistics companies, one of which is being able to deliver products quickly and at low cost, which reflects directly in the way of sorting packages, needing to eliminate steps such as storage and batch creation. Our work presents a multi-agent system that uses trajectory data mining techniques to extract territorial patterns and use them in the dynamic creation of last-mile routes. The problem can be modeled as a Dynamic Capacitated Vehicle Routing Problem (VRP) with Stochastic Customer, being therefore NP-HARD, what makes its implementation unfeasible for many packages. The work's main contribution is to solve this problem only depending on the Warehouse system configurations and not on the number of packages processed, which is appropriate for Big Data scenarios commonly present in the delivery of e-commerce products. Computational experiments were conducted for single and multi depot instances. Due to its probabilistic nature, the proposed approach presented slightly lower performances when compared to the static VRP algorithm. However, the operational gains that our solution provides making it very attractive for situations in which the routes must be set dynamically. 

\end{abstract}

\begin{keyword}
E-commerce Logistics, Dynamic Capacitated Vehicle Routing Problem with Stochastic Customer, Multi-agent Systems, Data Mining, Big Data.  
\end{keyword}

\end{frontmatter}


\section{Introduction}

The increase in internet penetration and the easy access to online payment forms have created a favorable environment for e-commerce sales growth [\cite{ecommercefoundation}]. However, this new market has also brought significant challenges in delivery logistics, particularly for the recent demand of low-cost same-day deliveries.

Brazil saw an exponential increase in e-commerce sales in 2019, especially in electronic products, home, decoration, health, cosmetics, and perfumery products [\cite{webshoppers2018}]. The country has more than 200 million inhabitants, and a wide  territorial extension (larger than 8.5 million km$^2$) with big social-economical differences among regions. The percentage of e-shoppers population in 2019 was $50.6\%$, having more than $80\% $ of them concentrated in the south and south-east regions [\cite{ecommercefoundation}]. Delivery companies usually specialize in these regions, making express deliveries to remote places, far from large cities, which last more than 6 days with costs greater than 20 dollars [\cite{correiosBrasil}].

Loggi is an express delivery logistics company based on a shared economy model just like  Uber, Rappi, and Glovo. Independent drivers use Loggi's online platform to accept and make last-mile deliveries with their own vehicles. Loggi's biggest challenge is to create a gig economy based sustainable logistics network for same-day deliveries in the whole country.

\begin{figure}[ht]
    \centering
    \includegraphics[width=0.7\textwidth]{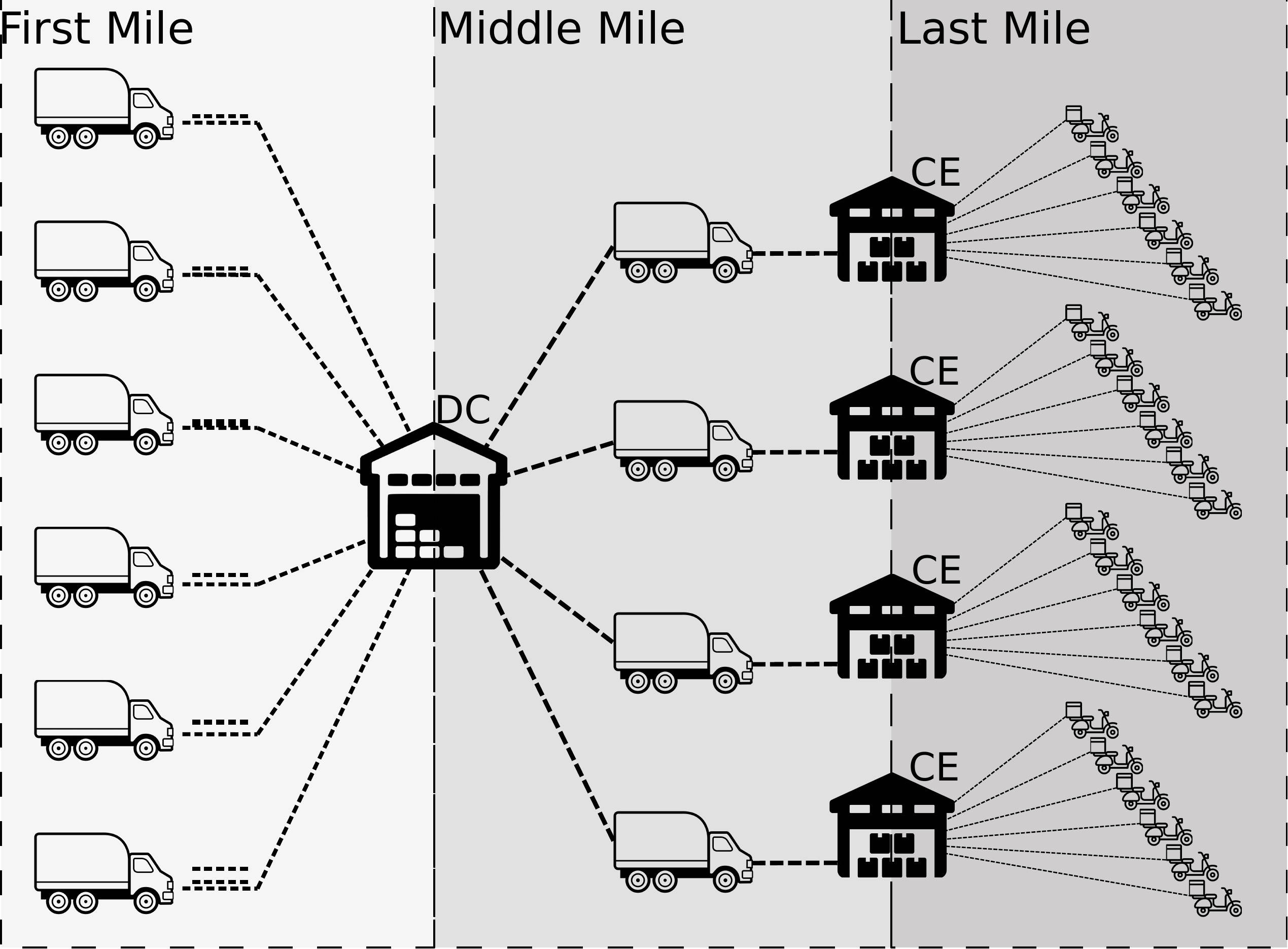}
    \caption{Delivery model used at Loggi.}
    \label{figModelo}
\end{figure}

The e-commerce delivery model used at Loggi is represented in Figure \ref{figModelo}, which is divided into three stages: first-mile, middle-mile, and last-mile. During first-mile, packages are collected in the e-commerce companies' distribution centers and sent to Loggi's distribution center (DC). In Loggi's DC, packages are grouped into unit loads based on distinct criteria such as priority, dangerous goods, and target Expedition Centers (EC). During middle-mile, packages are transferred from the DC to several expedition centers (ECs). This transfer is performed by trucks or planes, depending on the target EC location.
In this stage, the decisions of cutting or dispatch are made, a process known in the literature as waves between DCs and CEs [\cite{klapp2018dynamic}]. Finally, during last-mile, packages grouped into unit loads corresponding to delivery routes are collected by the independent drivers in the ECs, using cars, vans or motorcycles, and delivered to the final costumers.

To reduce the processing time and complexity of the package processing in the ECs, Loggi performs package routing in the DC. Packages are grouped into unit loads corresponding to delivery routes for all cities served by Loggi network in the DC, and  transferred to the ECs. The problem of creating last-mile routes is known as the  Vehicle Routing Problem (VRP)  [\cite{laporte1992vehicle}].

The VRP is usually solved by a static optimization algorithm, i.e., all information about the packages to be delivered (destination location, volume, weight, and so forth) are required prior to the beginning of its execution. In practice, this would mean that to use a static optimization algorithm for routing packages at Loggi's DC, it would be necessary to store all packages to be routed, run the optimization algorithm, and then split them in last-mile routes. However, given the number of packages processed at the DC per day and the company goal to perform same-day deliveries, storing everything in DC is unfeasible. It would require a lot of storage space and would increase the time and complexity of package's processing.

To solve this issue, Loggi demanded a dynamic VRP algorithm [\cite{ritzinger2016survey}], i.e., an incremental optimization algorithm able to process packages one-at-a-time, not requiring any storage.

The routing solution used by Loggi is formally described considering the following constraints: 

\begin{itemize}
    \item The vehicles used for last-mile deliveries have a maximum homogeneous capacity, considering weight, volume, and quantity of packages. This problem is known as Capacitated VRP (CVRP) [\cite{laporte2009fifty}].
    \item The company hires independent last-mile drivers and, for this reason, the routes created do not have to go back to the origin. This is known as Open VRP (OVRP) [\cite{schrage1981formulation,li2007open}].
    \item Routes are created in the DC and dispatched in ECs. This is known as Multi-depot VRP (VRPMD) [\cite{baldacci2013exact}] or Two-Echelon vehicle routing problem (2E-VRP) [\cite{crainic2009models,hemmelmayr2012adaptive}].
    \item Packages can't be stored in the DC. In other words, the algorithm is an Online VRP [\cite{jaillet2008online}]. Routes are created incrementally as the packages arrive in the DC. This problem can be defined as a dynamic VRP (DVRP) [\cite{psaraftis1988dynamic,pillac2013review}].
    \item Packages are unknown until they arrive in the DC. However, based on historical data, the package probability distribution by location can be calculated and used to build better routes. This represents stochastic information to consumers or VRP Stochastic Consumers [\cite{bent2004scenario,van2004dynamic,van2010online}].
    \item Some particular features of Brazil such as population, territorial extension, and number of cities to be attended (around 5700 cities), require the problem to be modeled as a Big Data problem.
 \end{itemize}
 
Some solutions have already been proposed for VRPs with the characteristics mentioned above. \cite{ausiello2001algorithms} presents the use of an online TSP algorithm (Traveling Salesman Problem) as a possible solution for same-day pickup and deliveries. In this implementation, routes are created incrementally using probabilistic TSP models, which takes advantage of historical information of deliveries already performed. The solution considers constraints such as vehicle capacity and open routes. However, this solution is not directly applicable to Loggi's problem since it only optimizes a single route, i.e., it solves the TSP and not the VRP. \cite{zhong2007territory} presents a model for e-commerce product delivery. The primary goal is to build realistic models to improve territory planning and vehicle dispatch process. By considering random customer locations and demands, and maintaining driver familiarity with their service areas, such solution provides flexible routes with more familiar drivers, resulting in better service with lower cost. \cite{huang2018designing} presents a proposal for a two-echelon logistics system for the delivery of e-commerce. It starts selecting a good location to be delivered in the urban area, a city distribution center. From there, packages are first transported to a satellite, from which they are delivered to their final destinations. They present a constrained delivery strategy using cells, and considers two formulations of the connectivity constraints, both inspired by multi-commodity flow concepts. These models can represent a viable solution to the problem here presented; nevertheless, these models are based on delivery to customers in densely populated urban areas, feature not present in all cities of Brazil.

The Multi-Agent (MA) architecture for Dynamic VRP \cite{thangiah2001agent,barbucha2009agent} has shown good results thanks to several features typically observed in multiple agent systems, like the autonomy of agents, the ability to increase computational efficiency through parallelization, and the possibility of using a distributed environment. In this case, each agent represents a unit load, they manage to solve problems such as capacity and closing unit load, independently. However, for a high degree of dynamism, the algorithm presented in \cite{barbucha2009agent} does not scale well. Besides, the tests were carried out for databases with a maximum of 200 consumers, preventing its implementation for the problem posed in this paper, which may have more hundreds of thousands of consumers daily. Most papers found in the literature are tested in small databases [\cite{bujel2018solving}]. To deal with large volumes of data, or Big Data, some articles have used Big Data techniques on VRP [\cite{bertsimas2019online,wang2016big,kytojoki2007efficient}].

When working with large and real databases, one of the techniques used to decrease the computational cost is to divide the area into fixed regions, which is known as Territory-Based VRP [\cite{wong1984vehicle, zhong2007territory}]. In the Territory-Based VRP, one route is a sequence of regions visited by the vehicle, which can be represented by the union of Spatio-temporal data [\cite{cao2005mining,fu2017mining,lv2019discovering}].Several works have focused on pattern recognition in space-time data applied in different applications such as travel time prediction [\cite{nakata2004mining}], forecasting gathering events [\cite{khezerlou2019forecasting}], anomaly detection in the GPS [\cite{shih2016personal}], route prediction system [\cite{chen2011personal}], and location prediction [\cite{wu2018location}]. Trajectory data mining [\cite{zheng2015trajectory}] is an emerging area that can bring applications and improvements to algorithms like VRP.

Given the aforementioned issues about the complexity of the routing problem faced by Loggi, this paper proposes a VRP algorithm based on a MA system that uses trajectory data mining techniques to decrease computational cost and allow the use of historical information of past routes to improve the algorithm performance. The proposed MA model is a VRP streaming algorithm, enabling the generation of routes without having to store packages for optimization, increasing the processing capacity by solving problems that in the literature are represented as Big Data. The MA system brings the characteristic of modeling each route as an agent. Each route is modeled with its own features, such as homogeneous capacity, and open routes. This allows us to escalate the problem because agents can specialize in different regions, converting our approach to solve problems of Multi-depot successfully. Also, this article uses a distributed version of the FP-Growth algorithm [\cite{li2008pfp}], which molds the mining problem as Big Data, enabling the use of distributed computing to mine large databases efficiently. Experiments were performed on Loggi delivery records. With the aim of comparing our approach with the classic model of the logistics companies, the results were compared to the static VRP. The remainder of the paper is organized as follows. We review relevant literature in Section 2. We formally define the proposed algorithm Stream Capacity Vehicle Routing Problem with Stochastic Customers based on Trajectory Data Mining and System MA in Section 3. We present the results of an extensive computational study comparing different delivery strategies in Section 4. We conclude with final remarks in Section 5.

\section{Literature review}

The problem exposed in this article has specific characteristics presented in the previous section. Particularly, it consists in a Fully Dynamic VRP (DRVP) with Stochastic consumers applied to Big Data problems. In this section, a literature review is presented  for the DVRP with Stochastic Customers and also for  trajectory data mining  techniques applied to Big Data problems.

\subsection{DVRP with Stochastic Customers}

The Vehicle Routing Problem (VRP) is one of the most important and studied combinatorial optimization problems. It has been studied for more than 60 years \cite{laporte1992vehicle,toth2002vehicle}. \cite{golden2008vehicle} defined VRP as: let a directed graph $G=(V,A)$, where $V= \{0, 1,... , n \}$ is the set of $n+1$ nodes and $A$ is the set of arcs. Node $0$ represents the depot, and the other corresponds to the $n$ customers. Each customer $i \in V$ requires a supply of $q_i$ units from the depot. A fleet of vehicles is stationed at the depot and is used to supply the customers. The vehicle fleet is composed by $m$ different vehicle types, with $M=\{1,2,... , m \}$. For each type $k \in M$, $m_k$ vehicles are available at the depot, each having a capacity $Q_k$ and a fixed cost $F_k$. In addition, for each arc $(i,j) \in A$ and for each vehicle type $k \in M$, a non-negative routing cost $c_{i,j}^k$ is given. VRP calculates optimal travels for visiting all consumers using the fleet of vehicles, or in other words, minimizing the distance traveled by the vehicles to visit all consumers, starting and ending in the depots. The travel or route is defined as a pair $(R,k)$, where $R=(i_1,i_2,...,i_{|R|})$, with $i_1=i_{|R|}=0$, is a simple circuit in G, and $k$ is the vehicle assigned to the route. 

 The VRP objective is to minimize the distance traveled by vehicles to deliver/collect products, and is widely adopted in transportation problems. However, transport companies have specific needs, such as delivery priorities, time windows, availability of resources, uncertainty in demand, time, or customers. These motifs have represented the need for variations of this problem in literature [39]. The more important variations  are Capacity VRP (CVRP) [\cite{laporte2009fifty}], VRP with Time Windows (VRPTW) [\cite{braysy2005vehiclea, braysy2005vehicleb}], Dynamic VRP [\cite{pillac2013review}], routing with cross-docking [\cite{liao2010vehicle}], among others.

In the dynamic VRP (DVRP) formulation, also referred to as real-time or online VRP, some input data is revealed through time [\cite{bernardo2018robust}]. Solutions for this problem in the literature are classified depending on the type of dynamic information, e.g., new customer requests, demands, service times, and travel times. However, in applications such as supplying supermarkets or service stations, there is a stochastic component to the problem, which is known in advance. The problem presented in this work is defined as a Stochastic Dynamic VRP (SDVRP) or, more specifically, DVRP with Stochastic Customers. The information about the packages is only known when they reach the DC. However, we can use historical data about delivered packages as a stochastic proxy for the incoming ones.

The SDVRP is divided into two stages. The first one is the planning phase, in which preprocessed decisions are built based on historical data. Following is the execution phase that uses the preprocessed decision and information of the events taking place to create online descriptions that are in charge of building the final routes  [\cite{bernardo2018robust, ritzinger2016survey}]. Works have implemented different methodologies in these stages: the most common methodology in the planning stage is to build a pool with all possible solutions, and use route selection algorithms in the execution stage. \cite{bent2004scenario} proposed a dynamic VRP time windows with stochastic customers, where the goal is to maximize the number of serviced customers. The planning phase uses a multiple scenario approach (MSA) that generates and solves scenarios which includes both static and dynamic requests. In the execution phase, the consensus function selects the most similar plan to the current pool of routes. \cite{gendreau1995exact} proposed a vehicle routing problem algorithm with stochastic demands and customers. The problem arises when, due to the uncertainty of consumers, it is not possible to plan routes. In the planning phase, a set of routes is designed. In a second stage, when the set of present customers is known, these routes are followed as planned by skipping the absent customers. They formulated the problem as a stochastic integer program and solved it by means of an integer L-shaped method. \cite{barbucha2009agent} based their proposal on a MA system. In the planning stage, this method used the sweep algorithm of Gillett and Miller [\cite{gillett1974heuristic}] to create initial routes; each route is an agent that represents a vehicle. In the execution stage, requests are evaluated by the agents that represent the vehicles in a dynamically changing environment. The main contribution of this work was the viability of using a MA system in VRP, managing to take advantage of several features, such as the autonomy of agents, the ability to increase computational efficiency through parallelization, and the possibility of using a distributed environment.

Other methodologies have been proposed to solve the SDVRP. \cite{huang2018designing} proposed an algorithm based on high density populated urban areas. Through a two-echelon logistics system, they introduced a flow optimization algorithm on blocks containing fixed areas pre-established in a city. The work is based on real cases of urban delivery systems in China, focused on delivery volume growth, and handle day-to-day delivery volume variations.

\subsection{Trajectory data mining}
 
A route or trajectory can be defined as sampling the displacement of an object through space-time. Formally, $p=\{x,y,t\} $ represents the Spatio-temporal location of an object, where $x$ and $y$ are the angles of latitude and longitude that represent the position on the Earth referenced with the Equator and the Greenwich Mean Time, respectively, and $t$ is the instant of time the sample was taken. Therefore, a trajectory is a sequence of location points, $T=\{p_1,p_1,...,p_n\}$, where $n$ is the number of points that describe the trajectory.

Trajectory data mining is the study of techniques used to search patterns in temporal space data. \cite{feng2016survey} defines four main tasks in data mining trajectory: clustering, classification, pattern mining, and knowledge discovery. Some of the most representative works in trajectory data mining are: \cite{agarwal2018subtrajectory} proposed sub-trajectory clustering, which objective is to capture parts or segments of trajectories shared in a database, assuming that each trajectory is a concatenation of a small set of paths, with possible intervals between them. \cite{da2016online} proposed an Online Clustering method for trajectories. The purpose of this method is to look for groups of trajectories, called micro-groups, that represent the relationship between the object and the movement. For this, the author stipulates that the object can appear, disappear, maintain or update its trajectories in a time window. \cite{sun2018identifying} proposes a trajectory classification model based on a multiclass SVM (Support Vector Machine). The author's objective is to identify the type of vehicle using data from the global positioning system (car, small, and large truck). \cite{qin2018spatio} developed a space-time mining model to discover habitual patterns in people using trajectories created by smartphones. These patterns have two aspects of information: What are the typical patterns of people's displacement? How much does their behavior vary from day to day? The main objective was to create a model to predict if a person lies in a zone, depending on time and day of the week. \cite{da2016framework} presented a framework to discover patterns of mobility and adaptation in a sub-trajectories data flow. An incremental algorithm is proposed to capture the evolution of micro-groups. They define a micro-group as a structure that represents the relationship between objects in motion.

Trajectories can be classified into active or passive, depending on how the data are stored [\cite{wu2018location,zheng2015trajectory}]. Active trajectories are a temporal union of points of interest (POI) such as stores, parks, restaurants, malls, cities, or countries. Such trajectories are usually generated by social networks, e.g., Facebook, Instagram, Twitter, etc. Passive trajectories are sequences of Spatio-temporal locations generated by sampling global positioning systems. Based on this classification, last-mile routes are passive trajectories, since a global positioning system on last-mile driver's smartphones generates them. However, if we consider the VRP output as a sequence of points, the locations of each package to be delivered, we can classify them as active routes, and each point will be defined as a POI. 

Studies developed in \cite{cao2005mining,mamoulis2004mining,fu2017mining,li2018position,lv2019discovering,khezerlou2019forecasting, chen2011personal} divide the target area as a set of cells that represent regions within a map. Thus, a route can be represented by sequences of POI, cells, or areas. In the problem addressed in this paper, the region where packages can be delivered is typically known. Therefore, all possible cells in which packages can be delivered are also already known, so that frequent itemset mining techniques can be applied by considering each item as a POI or cell visited. 

The most representative studies in trajectory data mining that represent trajectories in sets of POIs are the following: \cite{cao2005mining} studied the problem of discovering sequential patterns, which are routes frequently followed by the object. They defined two challenges, the fuzziness of locations in patterns, and the identification of non-explicit pattern instances. For this, they proposed a method that  transforms the original sequence into a list of sequence segments and detects frequent regions in a heuristic way. In this study, substring tree structure and improved Apriori technique were used. \cite{mamoulis2004mining} suggested a framework to analyze, manage, and query object movements. They divided the map into spatial regions and replaced the location of the object with these regions. In the case of multiple trajectories of moving objects (or time-series), these are typically concatenated to a single long sequence. Then, an algorithm that discovers frequent subsequences in a long sequence was applied. They proposed a mining algorithm for retrieving maximal periodic patterns, based on the adaptation of the Apriori technique for Mining Partial Periodic Patterns in Time Series Database presented in \cite{han1999efficient}. \cite{lee2009mining} proposed a graph-based mining (GBM) algorithm for mining the frequent trajectory patterns in a spatial-temporal database. This work creates mapping graph and trajectory information lists, and traverses the mapping graph in a depth-first search manner to mine all frequent trajectory patterns. The results presented in this article outperforms the Apriori-based and PrefixSpan-based methods, but it utilizes the adjacency property to reduce the search space. Such property can not be used in our problem because the packages in a route may be part of not adjacent regions. \cite{fu2017mining} proposed a technique for frequent route pattern mining. Firstly, trajectory partition, location extraction, data simplification, and common segment discovery are used to summarize trajectory data, convert these trajectories into common segment temporal sequences (STS), and generate 1-frequent itemsets. Then, a spatial-temporal adjacency relationship-based pattern mining algorithm is proposed. The authors also showed that sequence pattern mining algorithms for transaction data, e.g., association rules mining (Apriori, FP-Tree) and sequence pattern mining (GSP, PrefixSpan), can not be directly applied to summarized trajectory data. This is because these algorithms have high time complexity, and they do not consider the spatial contiguity and temporal continuity between items. However, for the proposed implementation of our work, it is not necessary to maintain the temporal continuity between the packages' delivery in the dynamic construction of the routes, considering the packages arriving at the distribution center do not have a temporal pattern. Therefore, these techniques can be used.

However, Apriori is an algorithm that uses breadth-first search, which makes the runtime increase exponentially with the increase in the number of itemsets. Additionally, Apriori scans all transactions for each iteration; therefore, due to memory consumption, its application becomes unfeasible for large databases, as in our problem. FP-Growth [\cite{han2000mining}] is a mining frequent patterns algorithm that uses depth-first search, whose runtime increases linearly with an increase in the number of itemsets. This algorithm uses a frequent pattern tree (FP-tree) structure, an extended prefix-tree structure for storing compressed, crucial information about frequent patterns for mining the complete set of frequent patterns by pattern fragment growth. FP-growth method is efficient and scalable for mining both long and short frequent patterns and is about an order of magnitude faster than the Apriori algorithm. \cite{li2008pfp} proposed a distributed version of the FP-Growth algorithm (PFP). PFP partitions computation in such a way that each machine executes an independent group of mining tasks. Such partitioning eliminates computational dependencies between machines, and thereby, communication between them. This algorithm was selected for trajectory data mining in our work.

\section{Methodology}

The proposed dynamic capacitated vehicle routing problem with stochastic customer proposed uses the following definitions:

\begin{itemize}
    \item Customer: the person who will receive the e-commerce package. The $i$th consumer corresponds to the $V_i$ node of the graph $G$ in the VRP definition. $V_i$ s defined as a geographical location with latitude ($lat_i$) and longitude ($lng_i$).

    \item Package ($p_i$): product to be delivered to the customer $i$. Each package carries information on the location $V_i$ of the corresponding consumer, and characteristics such as volume ($v_i$) and weight ($w_i$). 
    
    \item Route ($T_j$): a sequence of packages to be delivered to the corresponding customers. $T_j = \lbrace p_1, p_2, p_3, ..., p_q \rbrace$.
    
    \item Vehicle: transport used to perform a route. Each type of vehicle $k$ has some capacity constraints such as the maximum volume $Q_{kv}$, the maximum weight $Q_ {kw}$ and the maximum number of packages that can delivered $Q_{kp}$.
    
    \item Unit load ($U$):  a set of items grouped into shipping containers that can be moved easily with a pallet jack or forklift truck [\cite{laundrie1986unitizing}]. In our case, a unit load is a set of packages assigned to a route. The route will be performed by a vehicle $k$, so the unit load needs to respect the capacity constraints of the selected vehicle  ($Q_{kv}$, $Q_ {kw}$ and $Q_{kp}$).
    
    \item Warehouse system ($S$): hardware or process element (or a combination of both) that enables the sorting and intermediate storage of goods in between two successive stages of a supply chain [\cite{bartholdi2008warehouse, boysen2019warehouse}]. In our case, we can define a warehouse system as a limited set of unit loads  $S = \lbrace U_1, U_2, U_3, ..., U_s \rbrace$ and a sorting logic used to assign each new incoming package to one of the unit loads. When a unit reaches its maximum capacity (or reach any other \textit{closing rule}, such as a timeout), it is closed and replaced by a new unit load. The closed unit load is ready to be transferred to the next stage of the supply chain.
    
    \item Expedition center ($CE$): warehouse or depot corresponding the origin of last-mile routes. Each $CE$ has a warehouse system and geographical location, i.e., a latitude and a longitude.
    
 \end{itemize}

This work presents a stream based warehouse system sorting logic. The input of the algorithm is a stream of packages and the outputs are unit loads. The sorting decisions are performed incrementally, i.e., for each new incoming package the logic assigns it to a unit load. The proposed sorting logic is based on a Multi Agent system using the asynchronous team (A-Team) paradigm, introduced initially by Talukdar [\cite{talukdar1998asynchronous}]. This paradigm has already been applied to solve the Vehicle Routing Problem [\cite{barbucha2008multi,barbucha2009agent,thangiah2001agent}]. However, our proposal uses trajectory data mining techniques to extract stochastic information from the distribution of packages and thus improve the bet made by agents, maintaining features such as the parallel processing and independence of agents. In multiple $EC$ approaches (or multi-depot) a pre-separation of packages between $CEs$ is required, and consequently, each $CE$ has its MA system.

\begin{figure}[ht]
    \centering
    \includegraphics[width=0.6\textwidth]{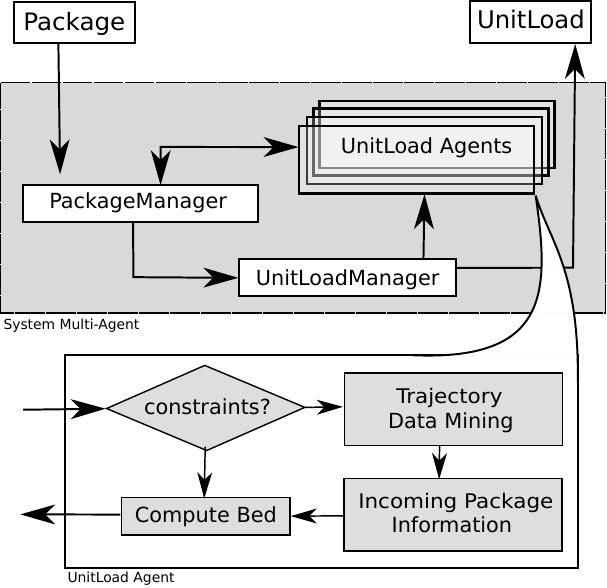}
    \caption{Structure of the MA.}
    \label{figMultiAgent}
\end{figure}

Figure \ref{figMultiAgent} presents the structure of the proposed MA system. For each new incoming package, the PackageManager agent collects the package information (destination, volume, and weight) and distributes them among all UnitLoad agents. Each UnitLoad agent is in charge of a unit load. The UnitLoad places a bet on each new package distributed by the PackageManager using the current state of the corresponding unit load. Finally, the PackageManager assigns the package to the unit load associated with the highest bet. Bets are estimated based on a convex combination of the similarity between the package and the packages already assigned to the unit load;; and some internal state of the UnitLoad agent representing  historical routes with a high likelihood of recurrence. The internal state of each UniLoad agent is computed before the sorting process starts. The unit load capacity constraints are verified for each new package, and if the eventual inclusion of a new package violates any constraint, the corresponding bet is 0. The UnitLoadManager agent receives information about the package and the unit load selected by the PackageManager. Then, it inserts the package into the unit load and decides whether or not to close it. If the unit load is closed, the UnitLoadManager runs a TSP algorithm using the set of packages assigned to the unit load and creates the corresponding last-mile route.  

The core function of the MA system is to efficiently compute the bet of a UnitLoad agent. The bet is defined as the convex combination of a value estimated via trajectory data mining ($bet_{DataMining}$) and other one estimated using only actual information of the packages in the unit load ($bet_{Distance}$), as defined bellow, where $\rho \in [0,1]$: 

\begin{equation} \label{eq:weighting}
bet = \rho  * bet_{DataMining} +(1 -\rho ) * bet_{Distance}
\end{equation}

\subsection{Trajectory Data Mining}

Association rules were introduced in \cite{agrawal1994fast} as a method for discovering interesting relations between variables in large databases, mainly used in products in large-scale transaction data recorded by point-of-sale systems in supermarkets. For example, the rule $\{\mathrm {meat,coal} \}\Rightarrow \{\mathrm {beer} \}$ found in the sales data of a supermarket  indicates that if a customer buys meat and coal together, they are likely to also buy beers. Such information can be used as the basis for decisions about marketing activities such as, promotional pricing or product placements. In correlation with trajectories, association rules methods applied to trajectories extract information such as $\{\mathrm {area_1,area_2} \}\Rightarrow \{\mathrm {area_3} \}$, which means that if a courier delivers packages in $area_1$ and $area_2 $, then he will deliver a package in $area_3$. The rule's antecedent is the area containing packages already in the unit load and the rule's consequent, areas containing packages that are likely to be part of the same route. This interpretation is the main assumption of this work, i.e., based on historical trajectory data it is possible to find relationships between packages that are used for the dynamic separation of routes. Besides, once the association rules are found, they can be stored in a hashtable data structure, where the key is the rule, and the output is the support, enabling a quick and efficient separation of packages.

The proposed sorting logic is divided into two phases. The first is the planning phase in which pre-processing and extraction of association rules is performed. Then, in the execution phase, in which the association rules extracted from historical data and real time information about new incoming packages are used to estimate the bet each UniLoad agent will put on a new package.

\subsubsection{Planning phase}

This phase is performed offline and needs to be updated periodically to adapt the extracted information due to consumer behavior changes or changes in business rules. The planning step is the task of creating association rules used for calculating the bet that each UnitLoad agent puts on new incoming packages. This phase is divided into two steps. The first step performs data pre-processing, where each historical route to be considered is transformed into a sequence of coarse grained geographic regions. In the second step the FP-Growth algorithm is used to build a FP-tree that represents frequent itemsets, i.e, frequent areas that are visited by historical routes. Association rules are extracted from the nodes of this tree.

We propose a pre-processing step that maps a sequence of points into a sequence of geographical areas named cells. This process is performed to reduce the granularity of the information to be processed by the FP-Grownth algorithm. The challenge is to represent the Earth in continuous computationally efficient areas for making queries. Google provides the S2 geometry library, which makes a spherical projection of the shape of the Earth, thus allowing us to map all points on the planet using perfect mathematics. Additionally, S2 is designed to have an excellent performance on large geographic datasets.  S2 geometry initially splits the Earth into six cells with an average area of 85011012.19 km2, and then  generates a tree of depth 30, where each level is a quarter of the cell area of the base node, being 0.74 cm2 the average area of the level 30 cells. In this way, the routes can be represented by areas of different levels, and each level defines different types of pattern extracted. Thus, the first step of the planning phase, maps each historical trajectory into a sequence of S2 cells, given a desired granularity adjusted by selecting an appropriate level of the S2 tree.

\begin{figure}[ht]
    \centering
    \includegraphics[width=0.5\textwidth]{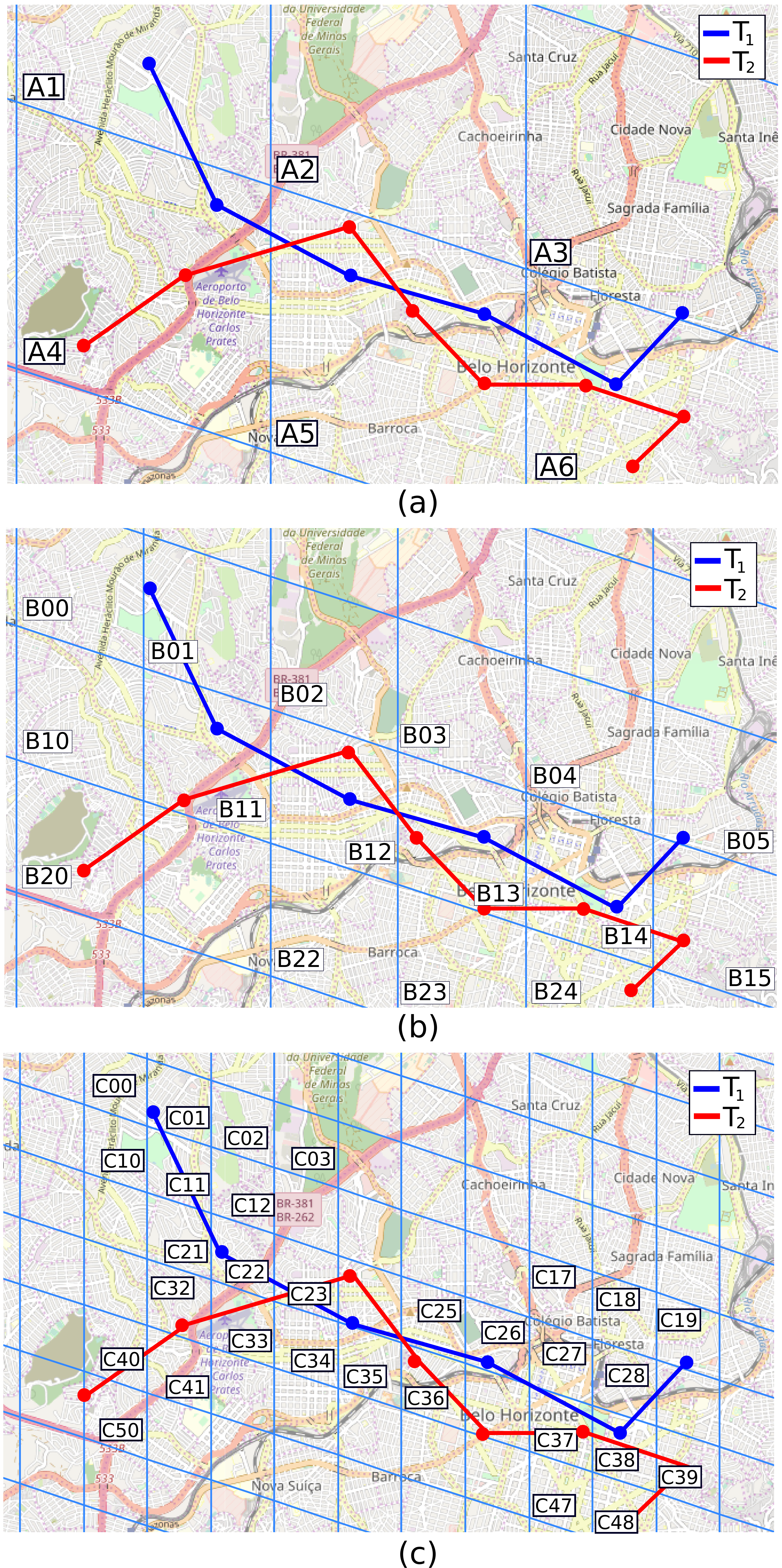}
    \caption{Pre-processing of the trajectories. Trajectory blue ($T_1$) and trajectory red ($T_2$) mapped in the S2 cells with different levels. Figure (a) with level 11 ($A$), (b) with level 12 ($B$), and (c) with level 13 ($C$). The mapped is represented in the Equation \ref{eq:map}.}
    \label{fig:Pre}
\end{figure}

Figure \ref{fig:Pre} shows an example of two trajectories ($T_1$ in blue and $T_2$ in red) mapped into S2 cells of different levels. This figure shows effect on adjusting the granularity of the information expressed by a trajectory using distinct levels of the S2 tree. The higher the level (a), the greater  is the  similarity between the routes, but the patterns extracted are more specific at lower levels. The resulting trajectories for each scenario are represented in the Equation \ref{eq:map}. For levels having larger areas, there is a larger similarity between the routes, since the majority of cells are the same, cells $A_4, A_5, A_6$ in Figure \ref{fig:Pre}a. Still, there is a loss of precision in the information presented, cell $C_{24}$ in Figure \ref{fig:Pre}c.  Cells of different levels maps trajectories into different patterns, so we recommend using different levels in the pre-processing trajectories.

\begin{equation} \label{eq:map}
    \begin{array}{l}
    T_1 = \lbrace p_{11}, p_{12}, p_{13}, p_{14}, p_{15}, p_{16} \rbrace\\
    T_2 = \lbrace p_{11}, p_{12}, p_{13}, p_{14}, p_{15}, p_{16}, p_{17}, p_{18} \rbrace\\
    \textrm{Level}_A: \left\{ \begin{array}{l}
    T_1 = \lbrace A_1, A_4, A_5, A_6, A_3 \rbrace\\
    T_2 = \lbrace A_4, A_5, A_6 \rbrace\\
    \end{array} \right.\\
    \textrm{Level}_B: \left\{ \begin{array}{l}
    T_1 = \lbrace B_{01}, B_{11}, B_{12}, B_{13}, B_{14}, B_{05} \rbrace\\
    T_2 = \lbrace B_{20}, B_{11}, B_{12}, B_{13}, B_{14}, B_{15}, B_{24} \rbrace\\
    \end{array} \right.\\
    \textrm{Level}_C: \left\{ \begin{array}{l}
    T_1 = \lbrace C_{01}, C_{22}, C_{24}, C_{26}, C_{28}, C_{19} \rbrace\\
    T_2 = \lbrace C_{50}, C_{32}, C_{24}, C_{36}, C_{37}, C_{39}, C_{48} \rbrace
    \end{array} \right.
    \end{array}
\end{equation}

The storage of large shopping databases (such as in supermarkets, websites, restaurants, etc.) has driven the development of techniques capable of predicting future purchases. Thus, grouping items to increase the likelihood of them being purchased. Data mining techniques are used to discover frequent itemsets. This problem is seen as mining association rules [\cite{leskovec2014mining}]. Similarly, in our problem, a trajectory can be modeled as a transaction (purchase made), where the visited cells can be represented as purchased items. Thus, the second step is implementing the FP-growth in the trajectories transformed into a sequence of cells (or items) and extracting association rules extraction on nodes of FP-tree built.

Fp-growth was introduced in \cite{han2000mining} as an algorithm for extracting association rules in databases containing transactions like databases of the routes. The same way that the Apriori uses the principle of monotonicity to carry out its search for association rules, ensuring that it does not have to sweep away all the possibilities of combinations of itemsets, which would have a prohibitive and exponential cost. FP-growth creates a compact tree structure called a frequent pattern tree or FP-tree, which moderates the multiscan problem and improves the candidate itemset generation.

To explain the implementation of FP-growth, we selected five trajectories and implemented the preprocessing step in S2 cells with level 12, presented in Equation \ref{eq:fp}. The goal is to find the set of frequent items at level 12, given these trajectories. The trajectories only visit the subset of cells $L_1 = \lbrace B_{01}:1, B_{02}:1, B_{05}:1 \rbrace$  of the total $ B $ cell in level 12. The first step of FP-growth is to count the number of paths that deliver packages in the $L_0$ cells (\ref{eq:fpL1}), to order in descending order (\ref{eq:L11}), and prune this vector using an input parameter called support, which  indicates how often it appears in the dataset, defined as

\begin{equation}
\textrm{Support(X)} = \frac{|X \subseteq t, t \in T|}{|T|},
\label{eq:support}
\end{equation}

\noindent where $t$ is the number of trajectories that visited the set of cells $X$, and $T$ is the total number of transactions. The support used in this example was $ 50 \% $, so items with less than three counts were removed.

\begin{equation} \label{eq:fp}
    \begin{array}{l}
    T_1 = \lbrace B_{01}, B_{11}, B_{12}, B_{13}, B_{14}, B_{05} \rbrace\\
    T_2 = \lbrace B_{20}, B_{11}, B_{12}, B_{13}, B_{14}, B_{15}, B_{24} \rbrace\\
    T_3 = \lbrace B_{20}, B_{21}, B_{22}, B_{23}, B_{13}, B_{14}, B_{15} \rbrace\\
    T_4 = \lbrace B_{10}, B_{11}, B_{12}, B_{02} \rbrace\\
    T_5 = \lbrace B_{12}, B_{13}, B_{14}, B_{15} \rbrace\\
    \end{array}
\end{equation}

\begin{equation} \label{eq:fpL1}
    \begin{array}{l}
    L_1: \left\{ \begin{array}{l}
    B_{01}:1, B_{02}:1, B_{05}:1, B_{10}:1, B_{11}:3, B_{12}:4,\\
    B_{13}:4, B_{14}:4, B_{15}:3, B_{20}:2, B_{21}:1, B_{22}:1,\\
    B_{23}:1, B_{24}:1\\
    \end{array} \right.\\
    \end{array}
\end{equation}

\begin{equation} \label{eq:L11}
    \begin{array}{l}
    L^{\prime}_1: \lbrace
    B_{12}:4, B_{13}:4, B_{14}:4, B_{11}:3, B_{15}:3 \rbrace.\\
    \end{array}
\end{equation}

\begin{figure}[ht]
    \centering
    \includegraphics[width=0.6\textwidth]{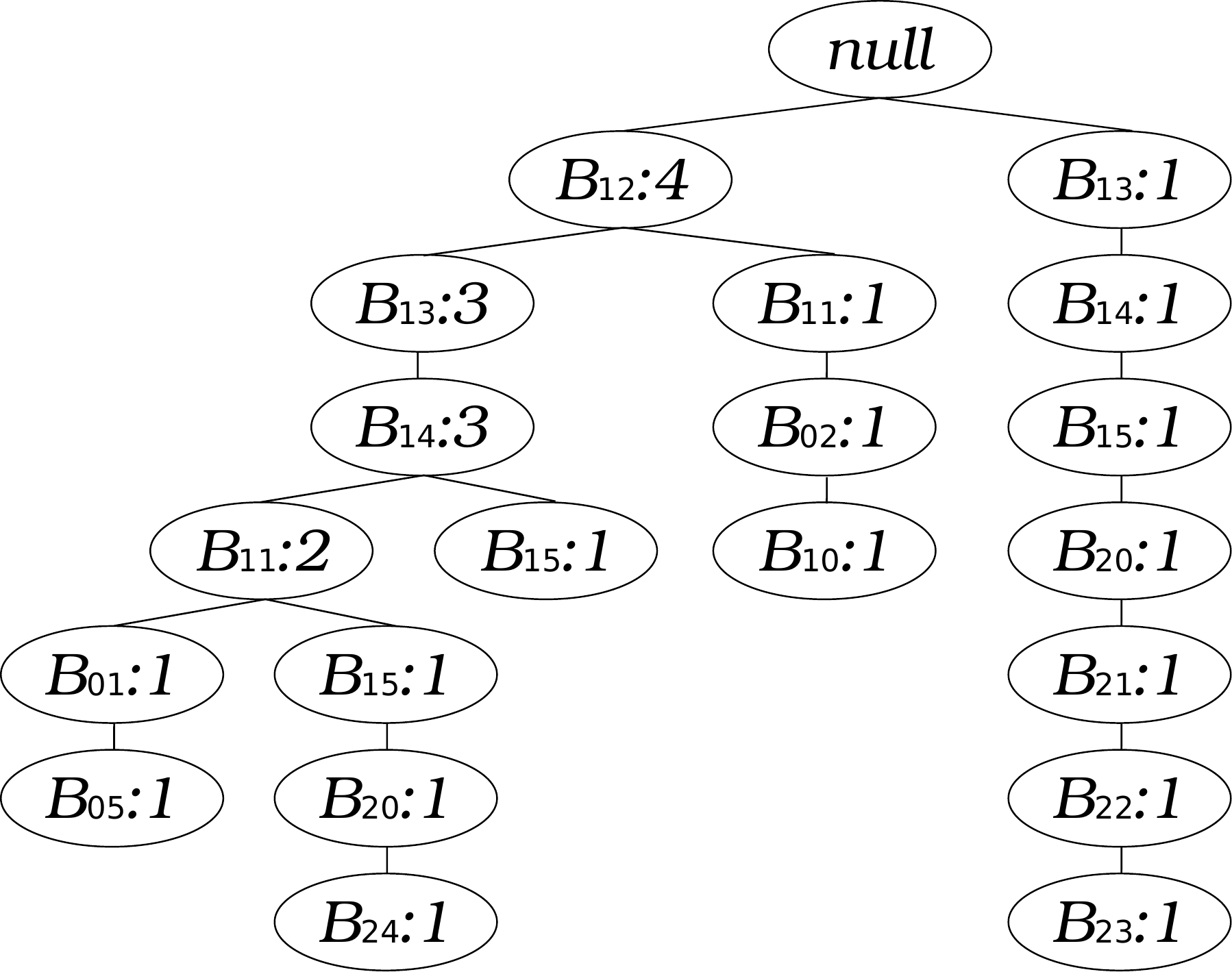}
    \caption{FP-tree built on trajectories \ref{eq:fp}.}
    \label{fig:tree}
\end{figure}

Given the set of frequent items $ L^{\prime}_1 $, the second is to build the FP-tree based on the set of trajectories given. The FP-tree is shown in Figure \ref{fig:tree}, its construction is done through each of the trajectories:

\begin{itemize}
    \item The items of $ T_1 $ are ordered based in $ L^{\prime}_1 $, thus generating $T^{\prime}_1= \lbrace B_{12}, B_{13}, B_{14}, B_{11}, B_{01} , B_{05} \rbrace $, where $ B_{12} $ is linked as a child to root, the root node is considered null. $B_{13} $ is linked to $ B_{12} $, $ B_{14} $ to $ B_{ 13} $, $ B_{11} $ to $ B_{14} $, $ B_{01} $ to $ B_{11} $, and $ B_{05} $ to $ B_{05} $, as shown in Figure \ref{fig:tree}. Each node is marked with the number of paths that pass through it, in the first instance all are marked with 1.
    
    \item Similarly to $T_1$, the trajectory $T_2$ is update in $T^{\prime}_2=\lbrace B_{12}, B_{13}, B_{14}, B_{11}, B_{15} , B_{20}, B_{24} \rbrace $. Since the $B_{12} $ to root connection already exists, the count on node $B_{12}$ is increased by 2 and we continue with the following connections. In this way, $B_{13}, B_{14}, B_{11}$ also has a count of 2 and $B_{15} , B_{20}, B_{24}$ are connected to the tree. $B_{15}$ to $ B_{11} $, $B_{20} $ to $ B_{15} $, and $ B_{24} $ to $ B_{20}$, these nodes are marked with 1.
    
    \item Likeness to $ T_1 $ and $T_2$, the trajectory $ T_3 $ is update in $ T^{\prime}_3 = \lbrace B_{13}, B_{14}, B_{15}, B_{20}, B_{21}, B_{22}, B_{23} \rbrace $. However, as in this trajectory the most frequent item isn't $B_12$, $B_{13} $ is linked as a child to root 2 the other nodes are connected in sequence, each one is marked with 1.
    
    \item The trajectories $T_4$ and $ T_5 $ were inserted in the tree in the same way as $ T_2 $ and $ T_3 $. So we built the FP-tree of Figure \ref{fig:tree} using $ T_2 $, $ T_2 $, $ T_2 $, $ T_2 $ and $ T_2 $, and support of the $50\%$. 
    
\end{itemize}

The last step is extract frequent items of the FP-tree based in the item of $L^{\prime}_1$, running this upward. The last item in $L^{\prime}_1$ is $B_{15}$, this occurs in 3 branch $\lbrace B_{12}, B_{13}, B_{14}, B_{11}, B_{15}: 1 \rbrace$, $\lbrace B_{12}, B_{13}, B_{14}, B_{15}: 1 \rbrace$, and $\lbrace B_{13}, B_{14}, B_{15}: 1 \rbrace $. Therefore considering $ B_{15}$ as suffix the prefix paths will be $\lbrace B_{12}, B_{13}, B_{14}, B_{11}: 1 \rbrace$, $\lbrace B_{12}, B_{13}, B_{14}: 1 \rbrace$, and $\lbrace B_{13}, B_{14}: 1 \rbrace$ . This forms the conditional pattern base. Immediately, we need to calculate the base conditional pattern considering the constructed FP-tree, which are the common prefixes with greater support for limiting support, for $B_{15}$ is $\langle B_{13}:3, B_{14}: 3 \rangle$. Finally, all combinations of frequent patterns are calculate. The process of extracting frequent items is shown in Table \ref{table:fpgrowthSummarized}.
 
\begin{table}[ht]
    \centering
    \caption{Mining of FP-tree summarized.}
    \label{table:fpgrowthSummarized}
    \begin{adjustbox}{max width=\textwidth}
        \begin{tabular}{c|l|l|l}
        Item                          & \multicolumn{1}{c|}{Conditional Pattern Base}                                                                                                                      & \multicolumn{1}{c|}{Conditional FP-tree}                                        & \multicolumn{1}{c}{Frequent Patterns Generated}                           \\ \hline
        \multirow{2}{*}{$B_{15}$}     & \multirow{2}{*}{$\lbrace \lbrace B_{12}, B_{13}, B_{14}, B_{11}: 1 \rbrace, \lbrace B_{12}, B_{13}, B_{14}: 1 \rbrace, \lbrace B_{13}, B_{14}: 1 \rbrace \rbrace$} & \multirow{2}{*}{$ \lbrace B_{13}, B_{14}: 3 \rbrace $}                          & $\lbrace B_{13}, B_{15}: 3 \rbrace, \lbrace B_{14}, B_{15}: 3 \rbrace,  $ \\
                                      &                                                                                                                                                                    &                                                                                 & $\lbrace B_{13}, B_{14},B_{15}: 3 \rbrace  $                              \\
        \multirow{2}{*}{$B_{14}$}     & \multirow{2}{*}{$\lbrace \lbrace B_{12}, B_{13}: 3 \rbrace, \lbrace B_{13}: 3 \rbrace \rbrace$}                                                                    & \multirow{2}{*}{$\lbrace B_{12}, B_{13}: 3 \rbrace, \lbrace B_{13}: 3 \rbrace$} & $\lbrace B_{12}, B_{14}: 3 \rbrace, \lbrace B_{13}, B_{14}: 4 \rbrace$ \\
                                      &                                                                                                                                                                    &                                                                                 & $\lbrace B_{13}, B_{14},B_{15}: 3 \rbrace  $                              \\
        \multicolumn{1}{l|}{$B_{13}$} & $\lbrace \lbrace B_{12}: 3 \rbrace \rbrace$                                                                                                                        & $\lbrace B_{12}: 3  \rbrace$                                                    & $\lbrace B_{12}, B_{13}: 3 \rbrace$                                      
        \end{tabular}
    \end{adjustbox}
\end{table}

The process of extracting association rules [\cite{piatetsky1991discovery,agrawal1993mining}] consists of finding, in a database, interesting relationships between two or more variables. For a better understanding, we utilize database $T$ set in Equation \ref{eq:fp}, each trajectory is mapped both S2-cells in levels 12 and is using FP-growth to extract itemset frequents represented in an FP-growth, exhibited in the Table \ref{table:fpgrowthSummarized}. Given an itemset frequent $\lbrace B_{13}, B_{14},B_{15} \rbrace$, can build the association rules: $B_{13}, B_{14} \rightarrow B_{15}$, $B_{13},B_{15} \rightarrow B_{14}$, and $B_{14},B_{15} \rightarrow B_{13}$.. In the first rule $B_{13}, B_{14}$ are the antecedent and $B_{15}$ is the consequent. 

However, these rules need to be validated using support (Equation \ref{eq:support}) and confidence. Confidence is the conditional probability that a transaction contains $Y$, since it contains $X$, defined as

\begin{equation}
\textrm{Confidence($X \mid Y$)} = \frac{\textrm{support}(X \cup Y)}{\textrm{support}(X)}.
\label{eq:confidence}
\end{equation}

To extract association rules in a dataset, regarding any two itemsets ($X$ and $Y$), one should create the desired rule and calculate the support of the itemsets involved. Once calculated, one can calculate the confidence of the rule using the Equation~\ref{eq:confidence}. Membership rules usually meet a minimum $\beta$ confidence level, and a minimum $\alpha$ support level, determined for each problem, according to the desired requirements. The problem of finding association rules is that the number of sets of items grows exponentially with the number of items in the database. Therefore, $\alpha$ and $\beta$ values define the depth of the mining algorithm and its computational cost.

\subsubsection{Execution phase}

In this phase, for each new incoming package to be assigned to a Unit Load, each UnitLoad agent computes a bet to be placed. The bet is defined by Equation \ref{eq:weighting} and is composed by two components. The first, $bet_{DataMining}$ is based on an association rule extracted from historical data and assigned to the agent during the planing phase. Given the features of the new package, the agent computes its association rule confidence using the already assigned packages as the antecedent of the rule. In other words, suppose we want to calculate the bet of the insertion of the point $p_{16}$ in the strajectory  
$T^{\prime}_1$ shown in Figure \ref{fig:Pre} and Equation \ref{eq:map}. First, the compute bid of the $T^{\prime}_1$ and $p_{16}$ in levels $A$, $B$, and $C$, are in \ref{eq:ex_dataMiningmap}, and then, the average of bet for different mappings defines the UnitLoad agent's bet. The average was used because it represents a weighting of the similarity between two routes. Taking up the previous example, $bet_{DataMining}$ as defined in Equation \ref{eq:bet_apriori}.

\begin{equation} 
\begin{array}{l}
bet_a = \textrm{Confidence($ A_1, A_4, A_5, A_6 \mid A_3$)}\\
bet_b = \textrm{Confidence($ B_{01}, B_{11}, B_{12}, B_{13}, B_{14} \mid B_{05}$)}\\
bet_c = \textrm{Confidence($ C_{01}, C_{22}, C_{24}, C_{26}, C_{28} \mid C_{19}$)}
\end{array}
\label{eq:ex_dataMiningmap}
\end{equation}

\noindent 
\begin{equation}
bet_{DataMining} = average(bet_A, bet_B, bet_C)
\label{eq:bet_apriori}
\end{equation}

\subsection{Distance information}

If the bet was based only the first component of Equation \ref{eq:weighting}, a new incoming package could not match any of the existing association rules, i.e., no agent would put a non-zero bet on it. One solution for this problem would be rejecting the package and present it to the system on a later time. However, this is not an efficient solution, since rejecting a large set of packages decreases operational efficiency. To avoid this scenario, as discussed previously, the bet is defined as a convex combination of two components:  one based on association rules and another one based on the minimum distance between the incoming package $p_p$ and all packages in each trajectory $T_v={p_1,p_2,...,p_n}$. This component is named  ($bet_{distance}$) and is defined as:

\begin{equation} 
\label{eq:bet_near}
    bet_{distance} = \frac{1}{1+ e^{\gamma (d_{min}(T_v) - \delta)}}
\end{equation}

Parameters $\gamma$ and $\delta$ of the logistic function can be found via cross-validation.  $d_{min}(T_v)$ is the minimal distance between the incoming package, $p_p$, and all other packages already assigned to a given trajectory $T_v$:

\begin{equation} 
\label{eq:d_min}
    d_{min}(T_v) = min(d(p_1,p_p),...,d(p_m,p_p))
\end{equation}

$d(p_x,p_y)$ is a metric that represents the distance between two points, e.g. Manhattan distance, great earth distance, or distance on road network \cite{aggarwal2015data}.

\section{Experimental Results} 

This section presents the experiments conducted to validate the effectiveness of the proposed approach. Initially, the chosen dataset is presented along with statistical analysis to support the hypothesis that there exist temporal correlations among spatial distributions of packages during weekdays. Next, the methodology for evaluating and configuring the models is described.

Our baseline for comparison is the use of the VRP static algorithm with different package batch sizes, representing a typical behavior of delivery companies, which store packages to set the routes. However, the proposed model is applied considering that the routes need to be generated as the packets arrive, dynamically. Hence, what we want to measure is how much our probabilistic data-driven approach differs from the baseline solution that has in advance all information required to solve the problem. As mentioned earlier, such a solution becomes infeasible when the volume of packages becomes very large.

\subsection{Dataset} 

The proposed routing algorithm was evaluated using a historical database provided by Loggi. The database corresponds to $112,000$ packages delivered in 2019 from 3 CE's located in the city of Belo Horizonte, Brazil. The dataset was split into a training and a test set. The training set has two distinct partitions: $100,000$ packages were used to build association rules in the trajectory data mining, and $1,500$ were used to select the hyperparameters $\gamma$ and $\delta$ of the proposed methodology. The test set is composed by $10,000$ packages. Each package is represented by its destination location (latitude and longitude), weight, volume, and the departing CE (depot). The routes were performed by motorcycles and are restricted by this vehicle's maximum capacity, i.e. a maximum volume of 110 liters, a maximum weight of 20 kg, and a maximum number of 25 packages. The warehouse system has 28 unit loads for each CE or origin; this number is based on Loggi's operational constraints. The closing of the unit load is performed when any of the amounts is equal or greater than $80\%$ of the limit of the vehicle.

\subsection{Statistical Assumption}

The Multi Agent system proposed in this work assumes that the geographical distribution of the package's destinations does not change significantly over a sequence of days. To confirm this hypothesis, we apply the Kernel Density-Based Global Two-Sample Comparison Test \cite{duong2012closed} using a historical dataset of delivered packages over one week. This is a non-parametric and asymptotically normal test proposed to compare cellular morphologies. The asymptotic normality bypasses the intensive computational calculations used by the usual resampling techniques to compute the p-value. Because all parameters required for the statistical test are estimated directly from the data, it does not require any subjective decisions. 

\begin{figure}[ht]
    \centering
    \includegraphics[width=0.7\textwidth]{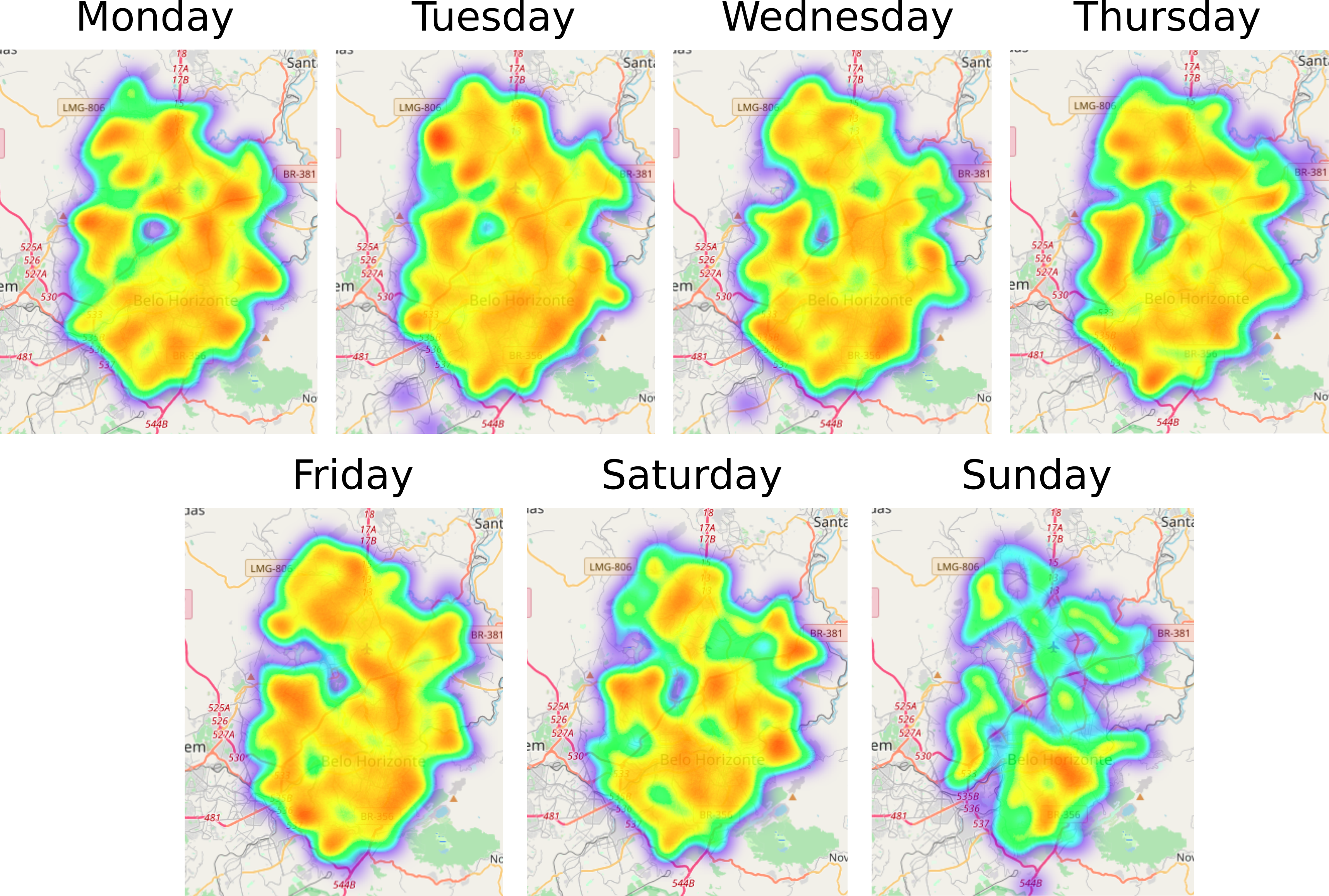}
    \caption{Heatmap of package distribution for a week in March 2019.}
    \label{figHeat}
\end{figure}

The test was implemented to verify whether the spatial distribution of the packages during weekdays in March was similar. The null hypothesis is  $H_0:F_l\equiv F_m$, where $F_l$, $F_m$ are the respective daily's densities of packages of a pair of days in a given week. Table \ref{table:pvalues} presents the p-values at a level of significance of $5\%$. 
As can be observed in Table \ref{table:pvalues}, there is no statistical evidence to reject the null hypothesis, that the spatial distribution of packages is similar over weekdays. Figure \ref{figHeat} shows the package's geographical distribution in the same week of March, illustrating graphically the results obtained in the test.

\begin{table}[ht]
    \centering
    \caption{p-value of kernel density based global two-sample comparison test for a week in March 2019.}
    \label{table:pvalues}
    \begin{tabular}{c|ccccccc}
    \hline\hline
              & Mon & Tue & Wed & Thu & Fri & Sat & Sun \\ \hline
    Mon   & 1      & 0.462   & 0.377     & 0.394    & 0.430  & 0.401    & 0.252  \\
    Tue   & 0.462  & 1       & 0.495     & 0.467    & 0.528  & 0.549     & 0.405  \\
    Wed   & 0.377  & 0.495   & 1         & 0.440    & 0.459  & 0.466    & 0.245  \\
    Thu   & 0.394  & 0.467   & 0.440     & 1        & 0.536  & 0.457    & 0.336  \\
    Fri   & 0.430  & 0.528   & 0.459     & 0.536    & 1      & 0.528    & 0.355  \\
    Sat   & 0.401  & 0.549   & 0.466     & 0.457    & 0.528  & 1        & 0.273  \\
    Sun   & 0.252  & 0.405   & 0.245     & 0.336    & 0.355  & 0.273    & 1    \\
    \hline\hline
    \end{tabular}
    
\end{table}

\subsection{Experimental Setup}

The experiments were implemented in Python using OR-tools [\cite{ortools}], which is an open source software suite for optimization, specialized in problems such as vehicle routing, flows, integer and linear programming, and constraint programming. The implementation of the FP-growth algorithm is based on PySpark. PySpark are the python bindings for Spark [\cite{spark2018apache}], an open-source framework for distributed computing, enabling the application of trajectory data mining in large data sets. We used the S2 library [\cite{s2Geometry}] of Google for selecting areas in the preprocessing phase of the trajectories.

The tuning of the hyperparameters $\gamma$ and $\delta$ of $bet_{distance}$ was performed using k-fold cross-validation [\cite{stone1974cross}].
Equation \ref{eq:cv} shows the range of values tested for these hyperparameters during the cross-validation process. The constants $\frac{10}{2000}$ and $5000$ are typical scaling factors for these parameters. 

\begin{equation} \label{eq:cv}
    \begin{array}{l}
    \gamma:\frac{10}{2000}*[0,0.1,...,0.9,1] \\ \\
    \delta:5000*[0,0.1,...,0.9,1] \\
    \end{array}
\end{equation}

As mentioned earlier, the proposed approach intends to generate the routes dynamically, assuming that packages arrive in a continuous flow. For evaluation purposes, its performance is compared with a VRP static algorithm with different batch sizes: 500, 1000, 2000, and 5000 packages. This baseline approach represents the typical behavior of a company that stores all packages before generating routes. In the case of Loggi's business model, this requirement is no longer feasible given the large volume of packages. Furthemore, the problem needs to be modeled with single and multi-depots. These two situations were tested using the same dataset and parameter setting. However, the trajectory mining phase is generated individually for each case (single and multi-depot) due to the association rules have different origins.

The extraction of association rules was performed over routes created using VRP static on package batches generated from the training subsets. For the single depot case, we considered a depot at downtown and divided the training packages into 20 batches of 2000 packages each. Thus, the set of 6006 created routes was mapped to areas of S2 cells of different sizes, as illustrated earlier in Figure \ref{fig:Pre}. 

Similarly, for the multi-depot case, the training packages were split into 3 depots located in the city and, with batches of 2000 packages. This procedure generates a set of 6434 routes, which was mapped to cells S2 and used to extract association rules. The Table \ref{table:features_singleDepot} shows the characteristics of cells S2 and the number of association rules extracted at each case and level. 

\begin{table}[ht]
    \centering
    \caption{Characteristics of cells s2 and the number of association rules extracted with FP-GROWTH at each level.}
    \label{table:features_singleDepot}
    \begin{tabular}{cc|cc}
    \hline\hline
    \multicolumn{2}{c|}{Cell} & \multicolumn{2}{c}{Association Rules} \\
    Level    & Average area (Km$^2$)   & Single-depot & Multi-depot \\ \hline
    11       & 20.27          & 691      & -  \\
    12       & 5.07           &  4541    & 3684 \\
    13       & 1.27           &  17844   & 25504 \\
    14       & 0.32           &  72080   & 87776 \\
    15       & 0.08           &  174147  & 160713 \\
    \hline\hline
    \end{tabular}
\end{table}

\subsection{Single-Depot} 

VRP is an optimization problem in which the objective is to minimize the distance ($d$) traveled by all vehicles. However, our problem is applied to Loggi's business model, and for drivers, a route with more packages is more attractive. Therefore, besides the travelled distance ($d$), other metrics analyzed in the experiment are the number of routes and packages per route. 

Results of the proposed model compared to the static implementation over batches of different sizes are presented in Table \ref{table:response_singleDepot}. The performance ratios with respect to the static VRP are showed in Table  \ref{table:response_singleDepot_comparative}. In general, It can be observed that the proposed method is slightly worse than the static VRP for all batch sizes. However, the implementation brings advantages to the dynamic application, avoiding the storage step, and creating cuts in the operationally ineffective packages. This approach also brings gains in running time, being more appropriate to Loggi’s model whose objective is to deliver cheaply and in the shortest time. 

When compared with the static VRP, the number of routes (or the number of vehicles used to deliver packages) exceeds up to a maximum of $17\%$; and the number of packages per route is reduced to about $15\%$ average. Still, these losses are expected to save time and operational cost. Furthermore, no package storage is required. It is also worth noting that the implemented process is dynamic, so that at the end of the day, it is necessary to force the closure of the unit loads. 

\begin{table}[ht]
    \centering
    \caption{Experimental results to single-depot problem}
    \label{table:response_singleDepot}
    \begin{tabular}{c|cccc}
    \hline\hline
    Model                       & Batch         & $Distance (km)$       & $Routes$   & $Packages$  \\ \hline
    Proposed                    & 1             & 7.60e+03  & 802       & 12.46          \\  \hline
    \multirow{4}{*}{VRP static} & 500           & 6.39e+03  & 722        & 13.85       \\
                                & 1000          & 4.88e+03  & 698        & 14.32       \\
                                & 2000          & 3.67e+03  & 690        & 14.49       \\
                                & 5000          & 2.60e+03	& 685        & 14.59        \\
                                & 10000         & 1.99e+03	& 683        & 14.64       \\
                                \hline\hline
    \end{tabular}
\end{table}

\begin{table}[ht]
    \centering
    \caption{Experimental results with respect to the static VRP to single-depot problem}
    \label{table:response_singleDepot_comparative}
    \begin{tabular}{c|cccc}
    \hline\hline
    Model                       & Batch         & $Distance$       & $Routes$   & $Packages$  \\ \hline
    \multirow{4}{*}{$\frac{\textrm{Proposed}}{\textrm{VRP static}}$} & 500           & 1.12             & 1.11        & 0.89      \\
                                & 1000          & 1.49             & 1.14        & 0.87       \\
                                & 2000          & 2.04             & 1.16        & 0.86       \\
                                & 5000          & 2.92	           & 1.17        & 0.85        \\
                                & 10000         & 3.82	           & 1.17        & 0.85        \\
                                \hline\hline
    \end{tabular}
\end{table}

\subsection{Multi-depot} 

The behavior of the results for the Multi-depot case is similar to those presented for the single-depot, achieving though slightly lower performances. This probably occurs because the package residue of unit loads is higher at the end of the day, once this implementation has about 2 times more unit loads than the single depot one. Another possible reason is that the coverage of the agencies is shorter, making the extraction of route spatial patterns more difficult.  

\begin{table}[ht]
    \centering
    \caption{Experimental results to multi-depot problem}
    \label{table:response_MultiDepot}
    \begin{tabular}{c|cccc}
    \hline\hline
    Model                       & Batch         & $Distance (km)$       & $Routes$   & $Packages$  \\ \hline
    Proposed                   & 1             & 6.10e+03  & 842        & 11.88          \\  \hline
    \multirow{4}{*}{VRP static} & 500           & 3.23e+03  & 722        & 13.85       \\
                                & 1000          & 2.89e+03  & 709        & 14.10       \\
                                & 2000          & 1.94e+03  & 696        & 14.37       \\
                                & 5000          & 1.60e+03	& 688        & 14.53        \\
                                \hline\hline
    \end{tabular}
\end{table}

\begin{table}[ht]
    \centering
    \caption{Experimental results in relation to the static VRP to multi-depot problem}
    \label{table:response_multidepot_comparative}
    \begin{tabular}{c|cccc}
    \hline\hline
    Model                       & Batch         & $Distance$       & $Routes$   & $Packages$  \\ \hline
    \multirow{4}{*}{$\frac{\textrm{Proposed}}{\textrm{VRP static}}$} & 500           & 1.88      & 1.16         & 0.86      \\
                                & 1000          & 2.11      & 1.18        & 0.84       \\
                                & 2000          & 3.14      & 1.21        & 0.83       \\
                                & 5000          & 3.81	    & 1.22        & 0.82        \\
                                \hline\hline
    \end{tabular}
\end{table}

\subsection{Discussion}

\begin{figure}[ht]
    \centering
    \includegraphics[width=0.68\textwidth]{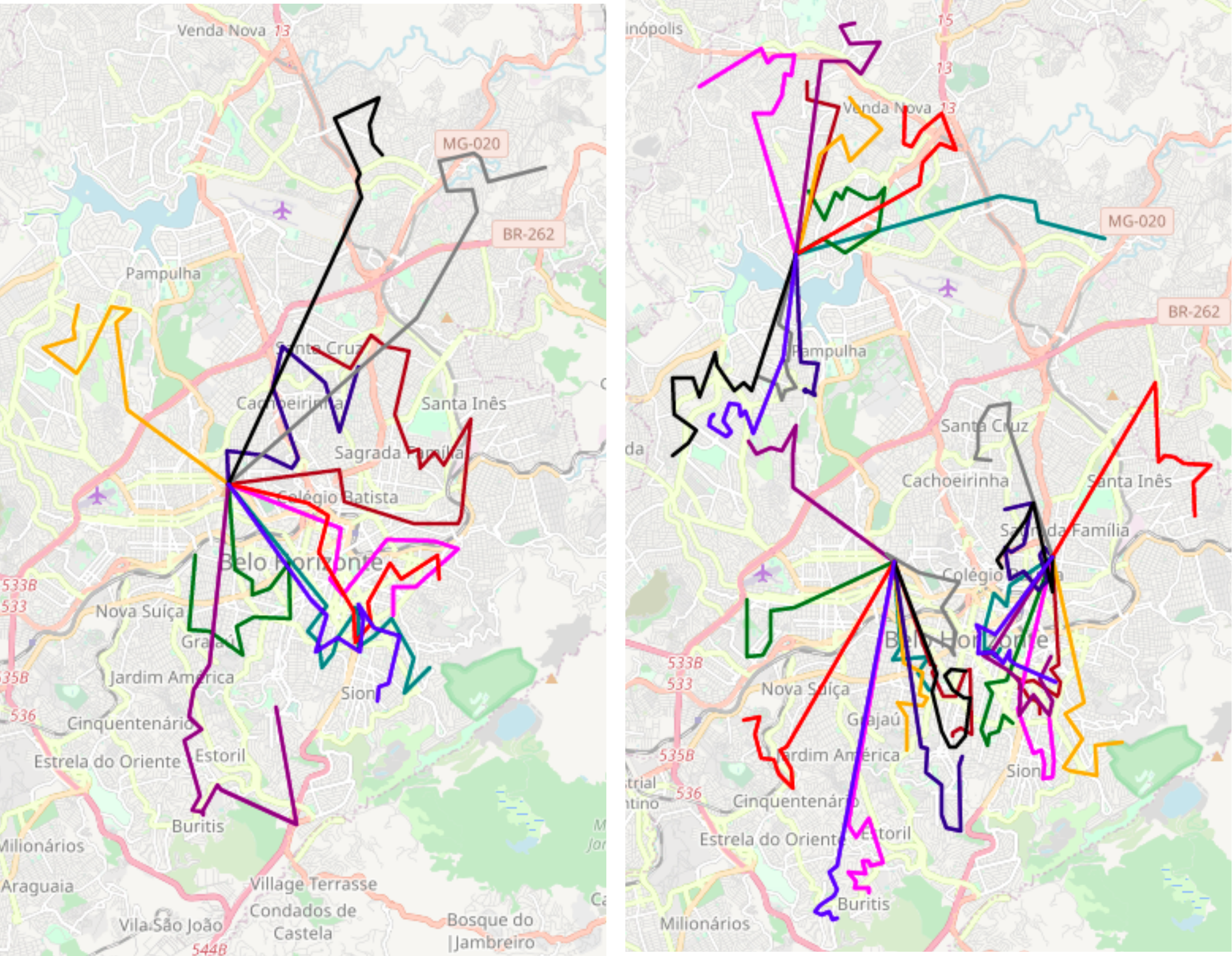}
    \caption{First routes created by the proposed model to each of the depots. The left map shows the first 10 routes of the single-depot approach, and the right map shows the first 10 routes of each CE for the multi-depot approach.}
    \label{figure:Routes}
\end{figure}

The experimental results demonstrates the viability of the proposed model as a possible solution for the Dynamic Capacitated Vehicle Routing Problem with stochastic customer. Still, it is not possible to analyze the patterns extracted on the trajectories. For this purpose, Figure \ref{figure:Routes} shows the first routes created by the proposed model for each of the depots. The left map shows the first 10 routes of the single-depot approach, and the right map shows the first 10 routes of each CE for the multi-depot approach. In both cases, the characteristics of spatial patterns can be observed. As expected, it were created specific routes to remote locations, joining packages from the same area, due to the implicit clustering of the routes represented in the S2 cells. The patterns used in the routing of the single-depot approach are those of larger coverage cells, since the same EC must supply the whole city; however, in the multi-depot approach, we can see the groups of packages more defined. This shows the effectiveness of trajectory mining techniques in extracting territorial patterns to produce coherent routes and satisfy the business problems in logistics's e-commerce, such as the dynamic creation of routes without having to store packages.

Our approach implements a CVRP with stochastic consumers, which allows the separation of packages using hardware already established in the industry, such as the use of the warehouse system. The results showed that the proposed algorithm creates more routes with a larger distance than the static model, but we understand that this comparison is not fair, since, in the static model, all the information on the problem is known in advance. A significant contribution presented in this work was to use data mining techniques to transform an NP-Hard problem into a linear problem based on the warehouse system's physical specifications. In other words, the complexity of our model is $\mathcal{O}(n_{CE}*n_U*(1+n_{Up}))$, where $n_{CE}$ is the number of CEs, $n_U$ is the number of unit loads of the warehouse system, $(1+n_{Up})$ is the computational cost of trajectory data mining algorithm and the near distance package inUnitLoadBed. Regarding this, since the association rules can be stored in a hash table and the compute bet can be calculated with $\mathcal{O}(1)$; and since the compute bet need to get the distance between a given package and all other packages at each unit load, thus $\mathcal{O}(n_{Up})$. Therefore, we conclude that the model's complexity does not depend on the number of packages processed. Accordingly, we can provide stability for problems with large amounts of packages (Big Data Problems), as is the case if the problem exposed in this work. Moreover, the process between CEs and UnitLoad agents are independent of each other and can be easily parallelized.

\section{Conclusions} 
 
This paper proposed a solution approach for the dynamic and stochastic CVRP focused on real e-commerce logistics problems. In this problem, the packages to be delivered arrive dynamically and need to be separated quickly without storage. From these requirements, we have modeled the problem as a continuous flow of input packages and output routes. Our solution is able to deal with Big Data scenarios, enabling the separation of a large number of packages efficiently. The problem was modeled on a warehouse system, where packages are dynamically divided between a fixed amount of unit loads. Unit loads are closed based on the operation criteria, as is commonly performed by logistics companies, especially e-commerce. Also, the problem was modeled as a multi-agent system that brings advantages such as the autonomy of agents, the ability to increase computational efficiency through parallelization, and the possibility of using a distributed environment. Each agent represents a unit load and manages to solve problems such as the capacity constraints and the need of closing unit load independently. The proposed algorithm uses territorial patterns extracted from trajectory data mining techniques to improve the bet made by agents, combining stochastic information to the model. The proposed method for extracting these patterns is an improvement of FP-growth, which enables the parallelization of the algorithm on distributed machines, allowing them to work efficiently with large volumes of data.

The results showed that the proposed method is an effective dynamic routing solution, presenting slightly worse performances when compared with the static routing solution based on package batch of fixed size. It is worth noting that these performance losses with respect to the static implementation were already expected due to the dynamism of the problem, and the operational limitations of the warehouse system. The losses for the multi-depot case were larger than compared with the single-depot ones. The possible reason is that the coverage area of each deposit is smaller and the association rules are not prone to extract these details. However, the multi-depot approach allows horizontal growth of the package separation, showing an operational advantage in the reduction of processes and time. The increase in the number of routes and the reduction of packages per route demonstrated that the algorithm is conservative, and the routes generated are still attractive to drivers, being very appropriate for the business model exposed in this work. 

Future work may be focused on exploring other types of typical uncertainties of the problem, such as stochastic and dynamic travel time. In addition, we plan to identify different settings such as robustness in the number of agents, which directly impacts the quality of the routes and the closure of the unit loads; this would enable a better use of the vehicles' occupation. Moreover, we also intend to explore other patterns in the routes that can be used to improve the bet of agents.

Additionally, this job uses greedy selection police, and we believe that techniques such as reinforcement learning would help both in the selection of the route and in closing them.

\section*{Acknowledgment}
This work has been supported by the Brazilian agency CAPES.

\bibliography{bib}

\end{document}